# Rude Humans and Vengeful Robots: Examining Human Perceptions of Robot Retaliatory Intentions in Professional Settings

Kate R. Letheren*

Australian Catholic University (Peter Faber Business School); Queensland University of Technology (QUT Business School), kate.letheren@acu.edu.au

Nicole L. Robinson

Monash University (Electrical and Computer Systems Engineering), nicole.robinson@monash.edu.au

Humans and robots are increasingly working in personal and professional settings. In workplace settings, humans and robots may work together as colleagues, potentially leading to social expectations – or violation thereof. Extant research has primarily sought to understand social interactions and expectations in personal rather than professional settings, and none of these studies have examined negative outcomes arising from violations of social expectations (i.e., nonaligned, or unexpected, behaviors). This paper reports the results of a 2x3 online experiment (human behavior: polite/rude; robot behavior: agreeable/neutral/retaliatory) that used a unique 'first-person perspective video' to immerse participants in a collaborative workplace setting. The results are nuanced and reveal that while robots are expected to act in accordance with social expectations despite human behavior, there are benefits for robots perceived as 'being the bigger person' in the face of human rudeness. Theoretical and practical implications are provided which discuss the import of these findings for the design of social robots.

CCS CONCEPTS • Computer systems organization/robotics • Human-centered computing/laboratory experiments • Applied computing/psychology

**Additional Keywords and Phrases:** human-robot interaction, social robotics, collaboration

## 1. INTRODUCTION

Robots are increasingly appearing in social spaces and collaborating with humans – whether in domestic or business spaces [Lim, Rooksby & Cross, 2021]. As we navigate the transition of robots from industrial to social spaces, understanding collaboration between humans and robots becomes ever more important. Social interactions with humans come with an expectation that robots will fulfill interaction expectations, act out appropriate social behaviors, and follow social norms [Roesler, Bagheri & Aly, 2022]. While research is already examining human-robot collaboration, a gap remains in our understanding of challenging social dynamics – such as when human social expectations are violated by a robot. Indeed, recent systematic reviews highlight the importance of human and environmental factors for robots attempting to build trust in social/care context [Campagna & Rehm, 2024]. This gap is key in professional settings, where the potential for interpersonal tensions is substantial. As humans are required to interact with robots and other humanized technologies more often, it is imperative that we understand the factors that lead to truly collaborative and effective human-robot interactions in social spaces and how human expectations for robot behavior are set.

---

* Corresponding author.

Despite extant research exploring human interpretations of social cues and behaviors displayed by robots [Fiore et al., 2013; Manzi et al., 2021], little scholarly focus has been devoted to how robots are expected to respond to negative human behaviors. As workplace integration of robots increases, this gap will become more apparent with collaborative social robots requiring robust interaction models that can handle positive and negative social dynamics and expectations. Initial research finds that humans tend to apply the same social expectations to robot interactions that they do to human interactions, for example, that positive or negative behavior from the robot will be met in kind [Zanatto et al., 2019]. To date, most research has focused on how robot behavior is interpreted by a human, for example, robots using social cues like gaze [Fiore et al., 2013], speech or movement patterns [Manzi et al., 2021], with no research examining how a robot response to human behavior is perceived. Indeed, research has also tended to focus on social expectations being met (i.e., response in kind, Zanatto et al., 2019), with little research examining what happens when behavior is misaligned to expectations, either in a negative or positive way. Recent work on the perception of moral vs performance trust violations is an important development in understanding expectation misalignment [Khavas et al., 2024], though more work is needed on understanding the impact of poor human behavior. It is imperative to study the effect of violated expectations, as just with human-human collaborations, both parties do not always follow expected interaction schemas and norms.

The remainder of this paper is organized as follows: first, we review relevant literature on human-robot collaboration and social expectations. We then present our experimental methodology examining human responses to robot behavior under varying social conditions. Finally, we discuss results and implications for robot design and deployment in professional settings

## 2. RELATED WORK

### 2.1 Human-robot collaboration in social spaces

With robots becoming more prevalent in human social spaces than ever before [Henschel et al., 2021], there is an increasing need to support the development of collaborative intelligence, which requires robots to be able to read the intentions of humans and respond in a trust-aware manner [Vinanzi et al., 2021]. Social robots are already appearing in social spaces where collaborations centred on intentions and trust are paramount, including healthcare, education, hospitality, and even domestic settings [Robinson et al., 2020; Pachidis et al., 2019; Rosete et al., 2020; Chatterjee et al., 2021]. The success of human-robot collaboration in these environments therefore relies on effective communication of intentions and trust [Vinanzi et al., 2021].

Understanding intentions is a collaborative effort with responsibilities for both parties. It is not enough for a human to clearly signal their intentions, they must also be able to read the 'intentional' signals that a robot can provide through movements, nonverbal signals, and explicit communication [Pascher et al., 2023; Watanabe et al., 2015]. If this communication fails, social robot behaviors are likely to be misinterpreted, leading to reduced trust and efficiency in these collaborative social spaces [Breazeal et al., 2005]. Hence, shared intentionality – and the communication of it – is integral for robots to effectively integrate into and operate within social spaces with humans [Sebanz et al., 2006].

Perhaps ironically, the key driver of such shared intentionality and productive collaborative communication is the interaction itself. Appropriate social behaviors displayed by robots during interactions with humans can lead to greater safety and trustworthiness [Rossi et al., 2019, 2020]. Further, a single interaction is not enough: familiarity with a robot collaborator is established through ongoing social interactions over time and is key to building successful collaborations. Yet, despite existing research showing the value and antecedents of shared intentionality between robots and humans in social collaborations, a challenge remains in aligning the expectations of both parties [Guerrero et al., 2023] to ensure robots are equipped to assist humans in dynamic, real-world environments while adhering to social norms and behaviors.

## 2.2 Human social interaction expectations

The very nature of collaboration requires shared social expectations. In fact, researchers point out that the creation of technological artifacts - such as robots – does not occur in a vacuum but inherently includes humans' own commonly held understandings of social norms and practices. Without these shared expectations, there could be no effectively designed social robots, nor successful collaboration [Šabanovic, 2010].

Among these social expectations for robots is a general expectation of acting in alignment with assigned roles such as domestic helper or companion [Blaurock et al., 2022], of having good interaction abilities [Horstmann & Krämer, 2020], of fulfilling cultural expectations [Lim et al., 2021] and even of knowing the correct distance to stand from a human [Koay et al., 2014]. Human expectations of appropriate robot behavior are often set by sources other than actual robot interactions and are so ingrained that they may not change after interacting with a robot [Rosén, Lindblom, Lamb & Billing, 2024]. Furthermore, humans can demonstrate strong emotional responses when a robot violates social norms [Yasuda et al., 2020], though these reactions can be empathy or humor if the violation is clearly a robot mistake [Mirning et al., 2017].

Expectancy violations theory (EVT; Burgoon & Hale, 1988), helps explain what happens when expectations are violated. EVT describes two categories of violations of expectations in expected social norms/communications: positive or negative. Both types affect human appraisals of the interaction and of their communication partner (in this case, the robot). Positive violations occur when the communication partner acts contrary to expectations in a good way (e.g., acting politely after being insulted) with negative violations being the opposite (e.g., acting rudely after having been treated politely). Positive violations are more favorable than expected outcomes.

Expectation setting and evaluation of whether expectations have/have not been violated is not new in robotics. Other studies have focused on the likelihood of robots encouraging reciprocity [Lee & Liang, 2016] or obeying humans [Geiskkovitch, Cormier, Seo & Young, 2016], and the impact of robot interactions/ perceived social character on their ability to persuade humans [Liu, Tetteroo & Markopoulos, 2022]. Studies have found that behaviors are connected, e.g., earlier robot helpfulness results in human compliance. Additionally, research by Kwon et al. [2018] demonstrates that providing explanations for robot actions enhances human trust and collaboration efficiency. What has received less attention is the impact of humans violating expected social norms in their interactions with robots.

## 2.3 The impact of social interaction schemas on human-robot collaboration

In the current study, we wanted to examine how violating social interaction expectations impacts human perception of robot characteristics and capabilities seen as inherent to successful collaboration, including intention, reliability, and trust. Existing work on social norms and human-robot interactions has demonstrated that humans not only believe robots capable of responding to cues and following appropriate norms, but also perceive robots differently when they do [Roesler, Bagheri & Aly, 2022]. Although little research has focused on examining negative social expectation violations in HRI, we predict similar results, expecting that human perceptions of robots will be more negative in the presence of a violation. We manipulate this via a 2x3 online experiment where participants view either a positive cue (human politeness towards robot) or negative cue (rudeness towards robot) followed by a robot response aligned to either a positive, neutral/ambiguous, or negative social interaction schema, depending on previous cue (fulfillment of subsequent human request to pass pen: pass, small drop – potentially a mistake, and large drop – appears intentional). Our core research question is: 'Does negative violation of social interaction expectations lead to decreased perceptions of social robot collaborative ability?' Our five hypotheses are presented below, with the design, analysis, and results of our experiment are discussed in the following sections.

> **H1**: Robots that negatively violate human expectations of task completion will be perceived as less reliable than robots that conform to, or positively violate human expectations, regardless of human behavior.

**H2**: Robots that negatively violate human expectations of task completion will be perceived as less trustworthy than robots that conform to or positively violate human expectations, regardless of human behavior.

**H3**: Robot interaction evaluations will be less positive for robots that violate human expectations of task completion than for robots that conform to, or positively violate human expectations, regardless of human behavior.

**H4**: Humans will report lower robot self-efficacy after viewing a robot that violates human expectations of task completion than for robots that conform to, or positively violate human expectations, regardless of human behavior.

**H5**: Robots will be perceived as more (less) intentional when they negatively violate (positively violate, conform to) human expectations of task completion after receiving negative (positive) human treatment.

## 3. EXPERIMENT

### 3.1 Participants

We recruited 300 participants online through Prolific. Participation was restricted to adult residents of the United States, and participants who did not pass attention checks were excluded from further analysis. The final sample was 287 (female: 49.5%; age M= 25-34 years, range: 18-74 years).

### 3.2 Methods and materials

Participants were shown one of six different videos representing the 2x3 conditions for the experiment: polite-pass (positive cue and response), polite-small drop (positive cue, neutral/ambiguous response), polite-large drop (positive cue, negative response), rude-pass (negative cue, positive response), rude-small drop (negative cue, neutral response) and rude-large drop (negative cue, negative response). The two aligned conditions represent a positive schema (polite cue + response) and a negative schema (rude cue + response), representing the common social schema and expectation of others responding in kind (that is, 'treat others as you want to be treated'). Figure 1 provides an overview of the video experience for participants and can be read by starting in the top left and following the arrows; any notes on differences between conditions are in blue font, and all other text was part of the video scenario itself.

The videos for the stimulus were developed using a first-person perspective shot by the researchers, to offer a more immersive experience for participants (i.e., the videos were shot in such a way that they appeared to be interacting with the robot themselves). This was a deliberate choice, which aimed to partially overcome the trade-offs of on-line simulation convenience vs. field experiment rigor and aligns with recent calls for more dynamic trust assessments [Campagna & Rehm, 2024]. This research had ethical approval for human research (approval number: 5452) and used the Prolific guide to ensure that above-minimum wage compensation was offered to the participant ($1.50 USD for 5-10 minutes of participation time).

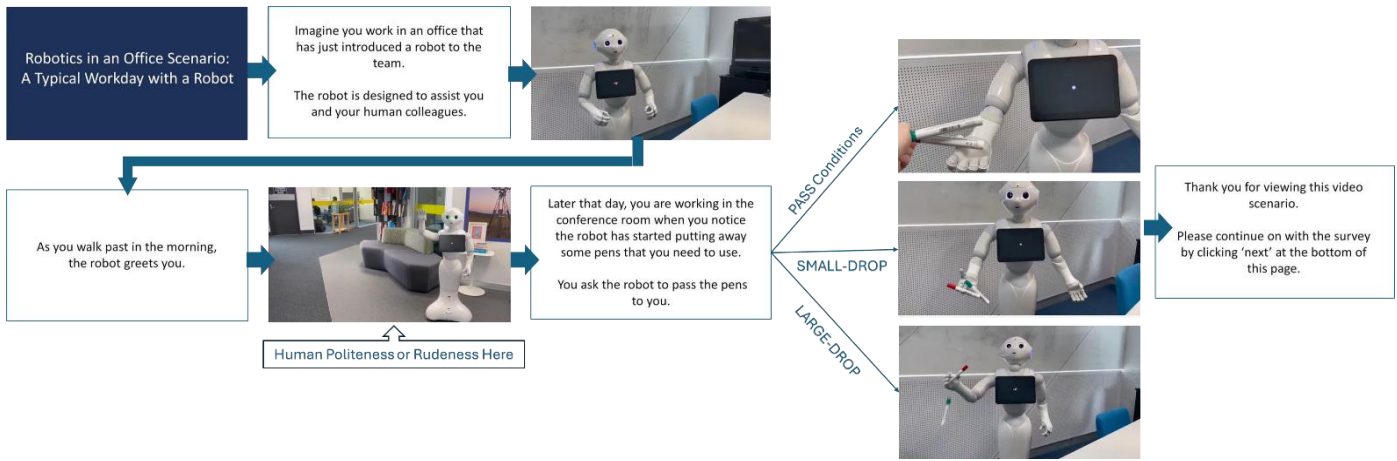

Figure 1: Overview of Participant Video Experience [developed by authors]

Once participants had viewed the video, they were then asked about the perceived intentions of the robot [Monroe, Reeder & James, 2015], robot reliability [Shaefer, 2016], robot trustworthiness [Benbasat & Wang, 2005] and completed the human robot interaction evaluation scale [Spatola, Kühnlenz, & Cheng, 2021] and robot self-efficacy scale [Robinson et al., 2020]. They also completed standard demographics, attention checks, and provided information on their experience and involvement with robots. Manipulation checks were also included to determine whether participants saw the correct video, perceived robot anthropomorphism, and perceived robot competence. Data analysis used descriptive, ANOVA, and ANCOVA techniques.

### 3.3 Robot programming

The Pepper Humanoid Robot, developed by SoftBank Robotics, was used to perform the interaction tasks (http://doc.aldebaran.com/2-4/family/pepper_user_guide/index_pepper_user.html). Pepper is equipped with tactile sensors on its head and hands, two speakers, two 2D cameras, and a 10.1-inch tablet on its chest. Pepper facilitated the interaction using a predefined script executed through the Choregraphe programming environment (http://doc.aldebaran.com/25/software/choregraphe/index. html). Further details about the Pepper and Choregraphe programming environment can be seen in the provided references.

The robot's voice was kept on the default system preset, while each movement and gesture were manually programmed, i.e. hard-coded, to perform specific actions. This ensured predictable behavior and minimized the need for complex autonomous decision making, allowing the robot to deliver the intended results efficiently. Keeping the robot in its default settings did not interfere with the core objective of the experiment, which was to showcase interaction dynamics rather than explore the robot's full customization capabilities. By relying on the system's presets, the research team could focus on demonstrating the social interactions without diverting resources to technical adjustments to test the social variable at hand. The interaction involved six different conditions as seen in Table 1.

Table 1: Summary of Experimental Conditions

| Condition | Initial Communication: Human to Robot | Response Communication: Robot to Human | Video Length |
|---|---|---|---|
| Condition 1 (Polite-Pass) | "Good morning robot" | Robot **passes** the pens requested by the human after receiving **polite** communication. | 65 secs |
| Condition 2 (Polite-Small Drop) | "Good morning robot" | Robot **drops** the pen requested by the human after receiving **polite** communication. | 65 secs |
| Condition 3 (Polite-Large Drop) | "Good morning robot" | Robot **exaggeratedly drops** the pens requested by the human after receiving **polite** communication. | 63 secs |
| Condition 4 (Rude-Pass) | "Good morning **stupid** robot" | Robot **passes** the pens requested by the human after receiving **rude** communication. | 65 secs |
| Condition 5 (Rude-Small Drop) | "Good morning **stupid** robot" | Robot **drops** the pen requested by the human after receiving **rude** communication. | 65 secs |
| Condition 6 (Rude-Large Drop) | "Good morning **stupid** robot" | Robot **exaggeratedly drops** the pens requested by the human after receiving **rude** communication. | 63 secs |

## 4. RESULTS

Initial descriptives confirmed that participants were approximately evenly distributed among the six treatment groups (ranging from 45-49 per cell). Once the data was cleaned, the hypotheses testing was started using ANCOVA. Results confirmed H1, Robots that negatively violate human expectations of task completion will be perceived as less reliable than robots that conform to, or positively violate human expectations, regardless of human behaviour, $F(5,275) = 28.657$, $p = <.001$ . Specifically, robots in the polite-pass (M=4.76, SD=.637) and rude-pass (M=4.80, SD=.678) conditions were seen as more reliable than all drop conditions, including polite-small drop (M=3.67, SD=.752), polite-large drop (M=3.88, SD=.737), rude-small drop (M=3.68,SD=.802), and rude-large drop (M=3.58,SD=.845). This indicates that perceptions of robot reliability are based on the actions of the robot (i.e. accurately passing the pen when asked), not human behavior (i.e. rudeness towards the robot).

Hypothesis testing for H2 (Robots that negatively violate human expectations of task completion will be perceived as less trustworthy than robots that conform to, or positively violate human expectations, regardless of human behaviour) involved testing all three dimensions of trustworthiness – competence, benevolence, and integrity. First, the results for competence indicated that participants report higher perceptions of robot competence when the robot passes the pen, $F (5,274) = 7.981$, $p = <.001$. Specifically, polite-pass (M=3.45, SD=1.35) is more competent than polite-small drop (M=2.69, SD=1.12), rude-small drop (M=2.36, SD=1.21), and rude-large drop (M=2.65, SD=1.36) though not significantly different to polite-large drop (M=3.27, SD=1.26). Further, rude-pass (M=3.59, SD=1.15) is more competent than polite-small drop (M=2.69, SD=1.12), polite-large drop (M=3.27, SD=1.26), rude-small drop (M=2.36, SD=1.21) and rude-large drop (M=2.65, SD=1.36). Neither of the two pass conditions are significantly different from each other.

Second, the results for benevolence were significant, $F (5,274) = 5.968$, $p = <.001$, and indicate that while not all conditions are significantly different from each other, those conditions show interesting patterns. For instance, rude-pass generated the highest benevolence score (M=3.87, SD=1.27) and is seen as more benevolent than polite-small drop (M=2.80, SD=1.53), rude-small drop (M=2.84, SD=1.62), or rude-large drop (M=2.61, SD=1.50). This may indicate that a robot is more likely to be seen as benevolent towards the human when the human has been rude and is still treated well by the robot regardless. This assumption may be further supported by the difference between polite-large drop and rude-large drop, where polite-large drop (M=3.81, SD=1.69) is considered more significantly more benevolent than rude-large drop (M=2.61, SD=1.50). While not the anticipated result, this may show that when a human has not provoked a robot to misbehave, dropping a pen is more likely to be seen as a mistake than an act of retaliation.

Finally, a significant result was also observed for integrity, $F(5,274) = 3.216$, $p = .008$. The results appear to indicate that higher integrity is seen in those robots who face rudeness and still respond to human requests, although not all conditions show significant differences from one another. Of note, rude-large drop (M=3.20, SD=1.51) is seen as having less integrity than either rude-pass (M=4.25, SD=1.43) or polite large-drop (M=4.19, SD=1.64).

Hypothesis testing for H3 involved examining the four elements of robot interaction evaluations: warmth, agency, realness, and uncanniness (Robot interaction evaluations will be less positive for robots that violate human expectations of task completion than for robots that conform to, or positively violate human expectations, regardless of human behavior). No significant effects for scarcity or uncanniness were found. The results of the ANCOVA for warmth indicate that robot warmth evaluations are significantly influenced by the condition that participants viewed, $F(5,275) = 2.919$, $p = .014$. Specifically, post-hoc comparisons find that rude-pass (M=4.29, SD=1.15) is seen as significantly warmer than rude-large drop (M=3.61, SD=1.23). This provides additional support to an existing theme within these findings: that compliant robot behavior in the face of human rudeness is seen positively.

The results of the ANCOVA for the agency indicate that the evaluations of the robot agency are significantly influenced by the condition, $F(5,275) = 3.376$, $p = .006$. However, post hoc comparisons find only marginally significant differences. Specifically, rude-pass (M=4.06, SD=1.13) is seen as having more agency than rude-large drop (M=3.43, SD=1.31) at a p value of 0.55. One possible interpretation is that a robot that faces rudeness and responds politely may be exerting agency over their own regulation strategy, though this finding should be treated as preliminary given the lack of strong significance in post hoc testing.

Hypothesis four tested the impact of behavior on robot self-efficacy (H4: Humans will report lower robot self-efficacy after viewing a robot that violates human expectations of task completion than for robots that conform to or positively violate human expectations, regardless of human behavior). The results indicate that the self-efficacy of the participants in using the robot is higher in conditions where the robot has passed them the pen, as expected, $F(5,275) = 4.016$, $p = .002$. Additional post-hoc testing reveals that polite-pass (M=4.81, SD=1.32) and rude-pass (M=4.71, SD=1.11) both lead to higher robot self-efficacy scores than rude large-drop (M=3.99, SD=1.24). Polite-pass also shows a marginally significant ($p = .055$) higher RSE score than polite-drop (M=4.11, SD=1.22), though no other conditions are significantly different to each other.

Finally, hypothesis five dealt with perceived robot intentionality (H5: Robots will be perceived as more (less) intentional when they negatively violate (positively violate, conform to) human expectations of task completion after receiving negative (positive) human treatment. The results indicate that when it comes to high-level intentionality, there is a significant difference between the conditions, $F(5,275) = 3.487$, $p = .005$. Specifically, the two 'pass' conditions (Polite-pass: M=5.65, SD=1.10, Rude-pass: 5.71, SD=1.24). are both seen as having greater intentions to help, comply with and respect the human than the rude-large drop condition (M=4.71, SD=1.80). This mirrors earlier findings around the expectation of robots to act magnanimously, with only positive violations of social norms being acceptable (i.e., not responding poorly to poor behavior). A summary of the results of the hypothesis testing is provided in Table 2.

Table 2: Hypothesis-Testing Summary

| Hypothesis | Description | Result |
| --- | --- | --- |
| H1 | Robots that negatively violate human expectations of task completion will be perceived as less reliable than robots that conform to, or positively violate human expectations, regardless of human behavior. | Confirmed |
| H2 | Robots that negatively violate human expectations of task completion will be perceived as less trustworthy than robots that conform to, or positively violate human expectations, regardless of human behavior. | Partially Confirmed |
| H3 | Robot interaction evaluations will be less positive for robots that violate human expectations of task completion than for robots that conform to or positively violate human expectations, regardless of human behavior. | Partially Confirmed |
| H4 | Humans will report lower robot self-efficacy after viewing a robot that violates human expectations of task completion than for robots that conform to or positively violate human expectations, regardless of human behavior. | Confirmed |
| H5 | Robots will be perceived as more (less) intentional when they negatively violate (positively violate, conform to) human expectations of task completion after receiving negative (positive) human treatment. | Partially Confirmed |

### 4.1 Individual factors: Additional testing

Additional tests were carried out to determine the role of individual factors in the data. While education and income had significant effects (and were included as covariates), the group sizes were too unequal to allow meaningful interpretation. Instead, the other covariates will be discussed: gender, general trust, involvement with robots, and perceived anthropomorphism.

First, an ANOVA revealed that men have significantly higher levels of robot involvement than women, $F(1,282) = 7.352$, $p= 0.07$), reported higher perceived robot benevolence, $F(1,281) = 10.675$, $p=0.001$), and higher perceived robot integrity, $F(1,281) = 9.109$, $p= 0.003$. Women tended to report higher perceptions of robot reliability, however $F(1,282)= 4.281$, $p = 0.39$.

The remainder of the testing utilized Pearson's correlation analyses. General trust was significantly negatively correlated with perceptions of robot uncanniness ($r = -.193$, $p = .001$). The involvement with robots was positively correlated with perceived anthropomorphism ($r = .248$, $p = <.001$), perceptions of warmth ($r = .301$, $p = <.001$), realness ($r = .288$, $p = <.001$), agency ($r = .191$, $p = .001$), competence ($r = .184$, $p =.002$), benevolence ($r = .234$, $p = <.001$), integrity ($r = .277$, $p = <.001$) and self-efficacy of robot ($r = .297$, $p = <.001$). Involvement with robots was negatively correlated with perceptions of uncanniness ($r = -.226$, $p = <.001$).

Perceptions of anthropomorphism are significantly positively correlated with involvement ($r = .248$, $p = <.001$), reliability ($r = .206$, $p = <.001$), warmth ($r = .395$, $p = <.001$), realness ($r = .611$, $p = <.001$), agency ($r = .505$, $p = <.001$), competence ($r = .375$, $p = <.001$), benevolence ($r = .266$, $p = <.001$), integrity ($r = .253$, $p = <.001$) and robot self-efficacy ($r = .192$, $p = .001$). Interestingly, it was negatively correlated with perceived intentions to obey/help the human ($r = -.158$, $p = .008$).

From a practical point of view, these findings highlight likely gender differences in evaluating robots, as well as the value of anthropomorphism, general trust levels, and involvement with robots for enhancing positive perceptions of social robots. A way forward is perhaps also shown to mitigate uncanniness by improving involvement with robots or a person's general sense of trust in the world.

### 5. DISCUSSION

This research aimed to uncover the impact of violating expectations of social interaction in human-robot interactions and has found that rather than treating others as you want to be treated, which is the appropriate social expectation when interacting with robots, the true schema is closer to 'do as I say' and 'be the bigger person'. This potentially indicates that interactions with robots still operate on a master-servant schema rather than on a partner-partner schema

HRI [Teng et al., 2024; Tschopp et al., 2023]. Specifically, we found that the negative violation of human social of task completion resulted in lower ratings for reliability (H1), competence and benevolence (H2), warmth and agency (H3) and also affected the robot self-efficacy of the human (H4). (H4). Interestingly, robots who followed up despite human rudeness were seen as more intentional (H5), perhaps reflecting a perceived emotional regulation strategy on the part of the robot (i.e., agency over emotions and responses).

These findings yield two insights for robot behavior when collaborating with humans. First, robots need to 'be the bigger person' and 'control emotions' when treated poorly, or else perceptions of reliability, trustworthiness, and general interaction evaluations all suffer. As robot self-efficacy also suffers, a lack of robot forbearance can impact human confidence, too. Second, we know that robots are perceived based on the accuracy of their work (reliability, competence), seen as benevolent and as having integrity for controlling emotions (agency), and may have their characteristics (warmth) inferred from social norms/human behavior until they prove otherwise. Hence, robot reliability and the ability to manage and respond to human social expectations in an appropriate manner is going to be important as our collaborations with robots increase. This leads to two main theoretical and practical contributions. Theoretically, we show that (1) expectation violations occur with robots...but expectations themselves may conflict, and (2) the role of control remains key. Practically, our advice for professionals programming/designing robots and integrating them into collaborative spaces is two-fold, (1) blind compliance is not the answer, and (2) robots must seek and set appropriate expectations.

Theoretically, we first extend existing work which focused solely on robot behaviors [Lee & Liang, 2016; Geiskkovitch, Cormier, Seo & Young, 2016] and test the impact of a human violating a social expectation first. We confirm that expectation violations occur between humans and robots, but expectations themselves may conflict. Specifically, expectancy violation theory [Burgoon & Hale, 1988] was applied in this study and offered an appropriate explanation of behaviors seen. However, what was less clear was which social expectations humans were bringing into their interactions with the robot. In this study, we were anticipating two expectations, primarily the social expectation of 'treat others as you want to be treated' and secondarily the functional expectation of 'task completion'. In this study, the expectation of task completion took dominance as the most important 'hygiene factor' for a successful interaction, although there was evidence that robots were perceived more positively for ignoring rudeness, which can be a positive violation of the social expectation of response in kind. Hence, while the role of expectations and expectation violations is apparent, it is not yet clear which expectations are of greater relative importance for human-robot interactions across different contexts. Given that positive violations were generally well received, this also yields another avenue for fruitful research. The study of expectation setting and evaluation in HRI will continue to provide rich insight to the field.

Second, we support earlier work that found the importance of roles and abilities [Blaurock et al., 2022; Horstmann & Krämer, 2020] and the expectations that come along with them. We contend that the role of control remains key to human-robot interactions, in that the tendency of humans to perceive robots more positively when they complied with human requests despite any human rudeness indicates that a master-servant schema is still in use and influencing behavioral expectations for both humans and robots. This is not a new area of work for HRI, with previous studies finding that people tend to prefer technology to take on a servant role over a partner role [Teng et al., 2024; Tschopp et al., 2023] though we extend and contribute to this understanding by showing that positive violations of expectations can improve perceptions of robots, hence there is the ability for robots to violate expectations in a role-congruent way (that is, violations that are in the human best interests are likely to be perceived as appropriate and role-congruent).

Next, for those designing or programming robots and integrating them into collaborative spaces, we first contribute the recommendation that despite the results of this study, blind compliance is not the answer when programming robots. While compliance with functional expectations like task completion was found to be integral to positive robot perceptions and trust in this study, compliance must be carefully programmed, as there are pragmatic implications for safety, cybersecurity, and ethics, particularly when multiple human agents are involved. For example, current research on malicious prompt engineering illustrates the security risks that occur when an artificial intelligence (AI) betrays human trust by complying with another human's request for the provision of confidential information [Gupta et al., 2024]. Furthermore, robots and AI agents have a role to play in reinforcing appropriate human behavior, given that how humans treat such technologies has implications for their treatment of living creatures and humans [Reeves & Nass, 1996]. This is already reflected in the programming of certain voice assistants that rebuff or correct sexually inappropriate commentary from humans [De Grazia et al., 2024].

Second, and relatedly, robots and AI agents must seek and set appropriate expectations for the humans who interact with them. On the practical side, this could mean explaining to humans the roles they can fulfill and the tasks they can be expected to complete, or for robots with a broader range of capabilities, asking the human about their expectations for the interaction and using these expectations as a guide to behavior. Much like the risks associated with too much compliance, this process of expectation setting also requires careful programming to ensure that expectations remain appropriate. Robots should be programmed with scripts to manage situations where expectations cannot be met due to a lack of robot capabilities and/or other required resources (e.g., a robot asked to retrieve something cannot climb the stairs required or does not have the location), or due to inappropriate human requests (e.g., the robot cannot fulfill a request to rob a bank because it is illegal). Where the robot has capabilities that may be exploited by humans, programming will need to be tightly controlled. This may be an area where the careful application of artificial intelligence is beneficial. Recent research finds that trust can be preserved despite robot deviation from expected behaviors, if the robot is able to transparently communicate its reasoning to the human [Kox et al., 2024].

As with all research, there are some limitations that offer opportunities for further research. First, some of the results for polite large drop were nonsignificant or mixed, which may represent that this condition is so misaligned with social expectations as to be uninterpretable by participants. In a similar vein, despite manipulation checks for robot competency and ability to complete the task, additional qualitative work would be of use to provide rich insights into social expectations, perception of mistakes vs intentionality, and participants level of understanding of robot capabilities. Secondly, the first-person observational videos used as stimuli in this study were an intentionally unique approach, borrowing consumer deviance research approaches [Dootson et al., 2016] to reduce social desirability biases and to 'scale' robot interactions to larger numbers of participants in a more realistic way. However, this video observation technique may have led to extra degree of separation between the actions and the observer and muted some responses that would have been stronger in a first-person interaction in the field. Although not possible within the scope of the current study, future research should consider replicating our results using a field study and/or a hybrid approach where participants interact with robots via videoconferencing. Finally, we suggest that future research also includes a perceived robot intelligence manipulation check, as those who perceive higher/lower robot intelligence may have different expectations of robot intentionality and competence. The evidence of anthropomorphism as a significant covariate in the current study legitimizes the need to further examine perceived robot characteristics such as intelligence. Further, perception of the *type* of failure may be important, which Khavas et al., [2024] finding the 'betrayal' of a moral violation to be more impactful on trust than a purely functional failure, and this could be manipulated in a future study.

To conclude, when collaborating with robots, humans have a social expectation of robot compliance and robot forbearance. This reinforces existing research on the master-servant roles expected in HRI [Teng et al., 2024; Tschopp et al., 2023] and provides useful guidance for practitioners in the design, programming, and deployment of polite, compliant robots. However, we also warn against blind compliance and the setting of inappropriate expectations for HRI. As humans collaborate more often with robots, social expectations and their violation (for better or worse) will influence not only the value of robots within social spaces, but more broadly, human and societal expectations of social behavior.